# Deep radiomic features from MRI scans predict survival outcome of recurrent glioblastoma

Ahmad Chaddad[1,3], Mingli Zhang[2], Christian Desrosiers[3] and Tamim Niazi[1]

[1] Department of Oncology, McGill University, Montreal, Canada
[2] Montreal Neurological Institute, McGill University, Montreal, Canada
[3] The laboratory for Imagery, Vision and Artificial Intelligence, ETS, Montreal, Canada
E-mail: ahmad.chaddad@mail.mcgill.ca

**Abstract.** This paper proposes to use deep radiomic features (DRFs) from a convolutional neural network (CNN) to model fine-grained texture signatures in the radiomic analysis of recurrent glioblastoma (rGBM). We use DRFs to predict survival of rGBM patients with preoperative T1-weighted post-contrast MR images (n=100). DRFs are extracted from regions of interest labelled by a radiation oncologist and used to compare between short-term and long-term survival patient groups. Random forest (RF) classification is employed to predict survival outcome (i.e., short or long survival), as well as to identify highly group-informative descriptors. Classification using DRFs results in an area under the ROC curve (AUC) of 89.15% (p<0.01) in predicting rGBM patient survival, compared to 78.07% (p<0.01) when using standard radiomic features (SRF). These results indicate the potential of DRFs as a prognostic marker for patients with rGBM.

## 1 Introduction

Gliomas are the most common type of primary brain tumor in adults. They can be classified by histolopathological features into four grades (I, II, III or IV) as mentioned in the World Health Organization (WHO). Grade I glioma correspond to non-invasive tumors, grade II/III to low/intermediate-grade gliomas, and grade IV to aggressive malignant tumors called glioblastoma (GBM) [1]. GBM is a devastating disease of the primary central nervous system with ubiquitously poor outcome and a median survival of less than 15 months [2]. Most patients relapse within months, after which there are limited options for further treatment [3]. Improvement of patient survival represents one of the biggest challenges for recurrent GBM (rGBM).

Radiomics analysis for the automated prognosis in brain tumor patients uses a wide range of imaging features computed from region of interest (ROI) as input to a classifier model [4, 5]. Standard radiomics approaches rely on a variety of hand-crafted features, for instance, based on histograms of intensity, grey level co-occurrence matrix (GLCM), neighborhood gray-tone difference matrix (NGTDM) and gray-level zone size matrix (GLZSM). In the last years, CNNs have achieved state-of-art performance for a wide range of image classification tasks [6]. A CNN is a multi-layered architecture that incorporates spatial context and weight sharing between pixels or voxels. Unlike standard radiomic techniques, which rely on hand-crafted features to encode images, CNNs basically learn image representations that are convenient for classification tasks, directly from training data [7]. The main components of CNN are stacks of different types of specialized layers (i.e., convolutional, activation, pooling, fully connected, softmax, etc.) that are interconnected and whose weights are trained using the back-propagation algorithm with some tuning functions (e.g., stochastic gradient descent with momentum). A common limitation of CNNs, when employed directly for prediction, is their requirement for large training sets which are often unavailable. An alternative strategy uses CNNs as a general technique to extract a reduced set of informative image features that are then fed to a standard classifier model. Since the CNN feature extractor and classifier are learned using separate training sets, this strategy is less prone to overfitting when data is limited. Recently, an approach using deep CNN features with a support vector machine (SVM) classifier was shown useful for predicting the survival of limited GBM patients [8]. Despite this success, the exploitation of multiscale features across different CNN layers as learnable texture descriptors remains limited. In this work, we argue that tumor progression can effectively be captured by texture descriptors learned using a CNN. Hence, we propose to extract deep radiomics features (DRFs) from a 3D-CNN with 41 texture quantifier functions, and use these features as input to a random forest (RF) model for predicting the survival of rGBM patients.

In our previous work [9, 10], a similar strategy was proposed for differentiating between normal brain aging and Alzheimer's disease (AD) in MRI data. Specifically, we used the entropy of convolutional feature maps as texture descriptors for classifying normal control versus AD subjects. In contrast, the current work considers a broader set of 41 quantifier functions to compute texture descriptors for predicting rGBM patient survival. We hypothesize that texture within CNN layers captures important characteristics of tumor heterogeneity which are highly-relevant for predicting clinical outcome (i.e. survival). Additionally, we address the problem of limited training data using a transfer learning strategy where the 3D-CNN to extract features is pretrained on MRI images for other prediction tasks.

## 2      Materials and methods

We describe our deep feature model DRF based on deep 3D CNNs and 41 standard radiomic features (SRFs), as shown in Figure 1. Post contrast T1-weighted images are first acquired for rGBM patients. Gross-total-resection (i.e., tumor ROI) are manually labelled in each scan using 3D Slicer tool. A set of 41 texture descriptors is then extracted from labelled images in two different ways: a) applying standard quantifier functions on the multiscale feature maps of a pretrained 3D CNN; b) applying these quantifier functions directly on the original ROIs. Various analyses are considered to evaluate the usefulness of DRFs to predict survival outcome. To identify features which are differentially enriched in short or long survivors, we separate patients based on the median value of radiomic features and assess survival difference using Kaplan-Meier estimator and log-rank test [11]. The 41 DRFs are used as input to a randrom forest (RF) classifier to separate rGBM patients into groups corresponding to short-term survival (i.e., below the median survival time) and long-term survival (i.e., above or equal to the median survival time). As a validation step, the statistical significance of resulted patient groupings is measured using the log-rank test and Kaplan-Meier estimator. All image processing, array calculations, significance tests and classifications were performed in MATLAB R2018b.

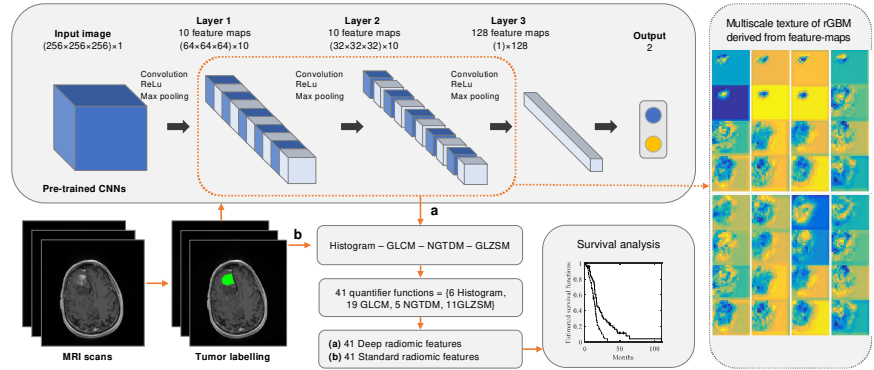

**Fig. 1.** Workflow of the proposed method to predict survival of rGBM patients. 1) identification and labelling GBM tumors in post contrast T1-WI MR images; 2) 41 quantifier functions encode the histogram, GLCM, NGTDM and GLZSM of feature maps (e.g. multiscale texture of rGBM) in layer 2 and 3 that derived from pretrained 3D CNN. These 41 DRF features in (a) compared to the 41 standard radiomic features using the log-rank, Kaplan-Meier estimator and RF classifier.

### 2.1 Datasets

This study uses a dataset of 100 rGBM patients with post-contrast T1-weighted (T1-WI) MR images. All images are derived from a unique site, and acquired using the same scanner model, pixel spacing and slice thickness. The volume datasets are resampled with a common voxel resolution of 1 mm$^3$, for a total size of 256×256×256 voxels. We normalized the intensities within each volume to a range of 256 gray levels. ROIs were manually labelled by experts using the 3D Slicer software 3.6. The labelling was performed slice by slice without prior clinical information.

**2.2 Proposed deep radiomic features (DRFs)**

To compute DRF, we used a pretrained 3D CNN architecture comprised of 4 layers, as shown in Figure 1. The 3D CNN architecture details are as follows. *Input*: image size = 256×256×256 voxels. *Layer 1*: filter size = 2×2×2; stride=2; filters=10; Max pooling; Rectified Linear Unit (ReLU); dropout=0.8; output = 10 feature maps of size (64×64×64). Layer 2: filter size = 2×2×2; stride=2; filters=10; Max pooling; ReLU; dropout=0.8; output = 10 feature maps of size (32×32×32). Layer 3: fully connected layer; output = vector size 128. Layer 4: softmax = vector size 2. The CNN was pre-trained on multi-site 3D MRI data for the classification of Alzheimer's[1], using cross-entropy loss, stochastic gradient descent optimization with momentum of 0.9 and learning rate of 0.0005. The main goal of using pretrained 3D CNN is to generate multiscale texture descriptors (Figure 1). These feature maps could be used directly as input to the classifier model (e.g., SVM, RF, etc.), however this leads to overfitting since the number of features largely exceeds the number of training samples. Instead, we apply conventional functions (e.g., Haralick's features [12]) to quantify the structure/texture within feature maps of deep CNN layers. Considering only the 3D feature maps in layer 1 (10 feature maps) and layer 2 (10 feature maps), we compute 41 DRF derived from the histogram, GLCM, NGTDM and GLZSM of 20 feature maps, respectively as following: 1) Histogram features (mean, variance, skewness, kurtosis, energy and entropy) encode the intensity level distribution for the image in the feature maps; 2) For GLCM, NGTDM and GLZSM features: image intensities were uniformly resampled to 32 grey-levels, averaging all the 19 GLCM features (angular second moment, contrast, correlation, sum of squares variance, homogeneity, sum average, sum variance, sum entropy, entropy, difference variance, difference entropy, information correlation$_1$, information correlation$_2$, autocorrelation, dissimilarity, cluster shape, cluster prominence, maximum probability and inverse difference [12]) across 52 GLCMs derived from 13 angles and 4 offsets, 5 NGTDM features (coarseness, contrast, busyness, complexity and texture strength [13]), and 11 GLZSM features (small zone size emphasis, large zone size emphasis, low gray-level zone emphasis, high gray-level zone emphasis, small zone / low gray emphasis, small zone / high gray emphasis, large zone / low gray emphasis, large zone / high gray emphasis, gray-level non-uniformity, zone size non-uniformity and zone size percentage [14]). Averaging the 41 features across the 20 feature maps is considered as the final descriptors, called deep radiomic features-DRF. Similarly, we computed the 41 SRF directly from original ROI images.

**2.3 Classifications and survival analysis**

To assess the proposed DRF in survival analysis, we considered days to death (i.e., censorship=1) or days to last visit (i.e., censorship=0) in uni- and multi-variate analyses. For the 41 DRF continuous features, the median value was used as threshold value to separate patients into two groups. For each group, the Kaplan-Meier method was con-

---

[1] https://www.github.com/hagaygarty/mdCNN

5sidered to describe the time-to-event distributions for each feature. The log-rank significance test was then employed to assess if either group was associated with the incidence of an event. An event/censor was defined as death or the last patient visit. To account for the multiple significance tests (41×2 variables), we corrected the p values Holm-Bonferroni method [15] and considered features with corrected $p < 0.05$ as significant. A multivariate analysis based on RF model was conducted on all the patient datasets (n=100).

Uncensored patients (n=6, alive patients) were included and were assigned the average survival time of the remaining patients with a time-to-death greater or equal to their own, as of the time at the last visit. Patients were then grouped into either short-term or long-term survivors. The threshold value dividing these two groups was their median survival of 14.88 months. We then used the combined 41 DRF as input for RF classifier model in a 5-fold cross-validation strategy. We used the RF classifier with 500 trees to predict the short- term and long-term survival outcome. The performance value is computed as the average AUC obtained across all 5 folds. To compare the DRF with the SRF we applied a similar 5-fold cross-validation strategy using the 41 radiomic features that were computed directly from the ROIs. To compare AUC value derived from DRF and SRF, we calculated significance using the chi-square test [16]. Importance values of the various features were computed within every RF tree, then averaged over the entire ensemble and normalized by dividing them by the ensemble's standard deviation. Positive importance values were considered predictive for an event.

## 3    Results

Figure 2a shows heatmaps of log-rank test p-values (negative $\log_{10}$ scale) for groups of patients divided by the median value of features. One feature derived from DRF (*High Gray-Level Zone Emphasis*) is associated ($p < 0.01$) with survival outcome of rGBM patients. In general, DRFs show a greater relationship to survival outcome than SRFs. As shown in Figure 2b, longer survival was associated with a lower *High Gray-Level Zone Emphasis* value (HR=2.28; CI=1.46-3.57; 13.36 vs. 16.45 months). Notably, *High Gray-Level Zone Emphasis* describes the heterogeneity texture of GBM tumor.

Assessment of the accuracy of RF models using the DRF or SRF as input features is done in Figure 2c. DRFs lead to a significantly higher accuracy ($p < 0.05$), with an average AUC of 89.15% compared to 78.07% using SRF, for predicting the short-term and long-term of survival outcome of rGBM patients. Comparing the AUCs of the predicted groups using the DRF and SRF, Chi-square test showed a significant p value < 0.0001. This result is consistent with the previous finding using the univariate analysis, in which DRF is more relevant than the SRF in predicting the survival. Once again, we applied the Kaplan-Meier estimator and log-rank test on the predicted groups (short-term and long-term survival) obtained by the RF classifier (Figure 5d). We observe that the patient groups obtained by DRF or SRF have significantly different survival outcomes with $p = 1.5 \times 10^{-6}$, HR=2.9, CI=1.82-4.7 and $p=6.8 \times 10^{-6}$, HR=2.96, CI=1.86-4.69, respectively.  To assess the importance of individual features, we combined the DRF and SRF to train the RF classifier model (Figure 2e). We find that the most predictive features (importance features > 0) are from DRF (i.e., Large Zone / Low Gray Emphasis).

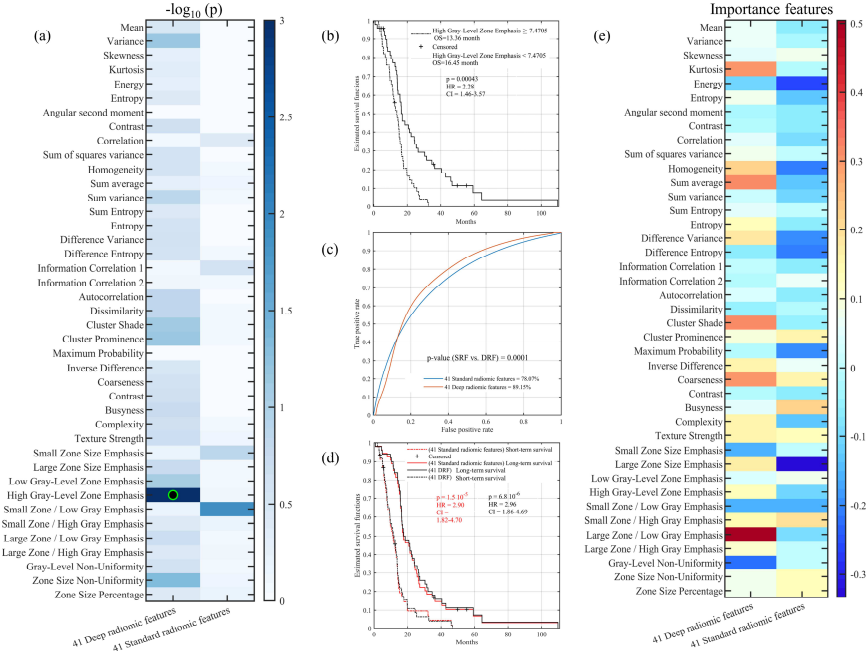

**Fig. 2.** Survival analysis using the deep radiomic features (DRF) and standard radiomic features (SRF). (a) Heatmap of log-rank test p values (negative log$_{10}$ scale) of survival difference between separated by individual features. Significant features (i.e., corrected p < 0.05) are indicated with a black-green circle. (b) Kaplan-Meier survival curves obtained for only significant feature (High Gray-Level Zone Emphasis). (c) The area under the ROC curves (average AUC on 5 folds) obtained by the RF classifier using DRF and SRF, for predicting patients with short-term (below-median) or long-term (above-median survival) survival outcome. (d) Kaplan-Meier curves of rGBM patients that significantly predicted by RF classifier model using DRF and SRF. Solid curves correspond to the long survival group and dot curves to the short survival group. (e) Importance of individual features for predicting the survival group with the RF classifier. Positive and negative values correspond to predictive and non-predictive features, respectively.

## 4      Discussion

Most models based on the radiomic analysis for GBM use SRFs which include histogram, texture, and shape features derived from MR images as a non-invasive means for predicting tasks [4, 5]. SRFs have been established as a technique for quantifying the heterogeneity related to tissue abnormalities [4, 5]. Furthermore, these studies have proven the link between imaging features and clinical outcomes. Deep multi-CNN channels corresponding to MRI modalities with SVM model have been recently demonstrated to effectively predict patient survival [8].

This study proposed an objective framework using DRF based on pretrained 3D CNN with RF classifier to predict the survival outcome of rGBM. We showed that the low value of deep *High Gray-Level Zone Emphasis* is associated with long-term survival. This feature describes the heterogeneity of active tumor of rGBM. The DRFs improve performance to predict the survival of rGBM with an average AUC of 89.15%, compared to 78.07% using the STF (Figure 2c).Our findings relate to previous studies, which found various radiomic features, in particular computed from ROIs of GBM, to be associated with overall survival [4, 5]. For example, texture descriptors derived from joint intensity matrix [4] have been shown to predict the prognosis of GBM patients. Likewise, radiomic subtypes defined by texture feature enrichment have been linked to differential survival [5]. Recently, fully-connected layer features derived from multi-channel 3D CNNs corresponding to multimodalities of MR images as input to SVM classifier were demonstrated to be prognostic for GBM. In this work, we report the potential of DRF in survival analysis. Notably, DRFs are encoded the information flow in CNN architectures. While, the impact of the information flow has been demonstrated previously in the computational biology field [17]. For example, a maximal information transduction estimation approach based on an information model was efficiently applied for transcriptome analyses [18]. Additionally, the findings from this study considered only 100 rGBM patients that requires a prospective validation on a larger dataset. Investigating additional DRFs by applying deeper architectures of CNN could potentially lead to better prediction of survival.

## 5    Conclusions

We proposed deep radiomic features derived from a 3D CNN to predict the survival of recurrent GBM patients. Our results show that these features lead to a higher classification performance than the standard radiomic features. Involving multi MRI modalities of rGBM tumor and more complex 3D CNN architectures to combine deep features with multi-omics (e.g., genetics + transcriptomics) could hold great promise for predicting clinical outcomes in rGBM patients.

8